\documentclass[12pt]{article}
\def\bSig\mathbf{\Sigma}

\usepackage{graphicx}
\usepackage{natbib}
\usepackage{multicol}
\usepackage{comment}
\usepackage{authblk}
\usepackage{mathrsfs,amsmath,amsthm,amssymb,color,amstext,mathtools}
\usepackage[colorlinks,breaklinks=true]{hyperref}
\usepackage{xcolor}
\usepackage{bm}
\hypersetup{
    colorlinks,
	linkcolor={magenta},
    filecolor={blue},
	citecolor={cyan},
}
\usepackage{enumitem}
\usepackage{breakcites}
\setitemize{noitemsep,topsep=0pt,parsep=0pt,partopsep=0pt}
\setenumerate{noitemsep,topsep=0pt,parsep=0pt,partopsep=0pt}
\usepackage[parfill]{parskip}
\definecolor{ao}{rgb}{0.0, 0.5, 0.0}
\newcommand{\St}{{\color{blue}\bm{S}_{t}}}

\newcommand{\Oini}{{\color{orange}\bm{O}_{1}}}
\newcommand{\Ot}{{\color{orange}\bm{O}_{t}}}
\newcommand{\Otn}{{\color{orange}\bm{O}_{t+1}}}
\newcommand{\Otq}{{\color{orange}\bm{O}_{t+q}}}
\newcommand{\Oi}{{\color{orange}\bm{O}_{i}}}
\newcommand{\Aini}{{\color{ao}\bm{A}_{1}}}
\newcommand{\At}{{\color{ao}\bm{A}_{t}}}
\newcommand{\Atn}{{\color{ao}\bm{A}_{t+1}}}
\newcommand{\Ai}{{\color{ao}\bm{A}_{i}}}
\newcommand{\Rt}{{\color{red}\bm{R}_{t}}}
\newcommand{\Ri}{{\color{red}\bm{R}_{i}}}

\newcommand{\blind}{1}

\usepackage{xcite}
\usepackage{styles}
\usepackage{nomencl}

\makeatletter
\newcommand*{\addFileDependency}[1]{
  \typeout{(#1)}
  \@addtofilelist{#1}
  \IfFileExists{#1}{}{\typeout{No file #1.}}
}
\makeatother

\begin{document}

\if1\blind
{
\title{\Large{\textbf{Statistical Inference in Reinforcement Learning: A Selective Survey}}}
\author[1]{Chengchun Shi\thanks{This paper has been submitted to the Festschrift in honor of Prof. Michael Kosorok, celebrating his remarkable achievements in the field of statistics and beyond. The author is deeply grateful for Michael's invaluable support and  guidance.}}
\affil[1]{London School of Economics and Political Science}
\date{\empty}
\date{}
\maketitle
} \fi

\if0\blind
{
\title{\Large{\textbf{Testing Nonstationary in Reinforcement Learning}}}
\author{
\bigskip
\vspace{0.5in}
}
\date{}
\maketitle
} \fi

\begin{abstract}
Reinforcement learning (RL) is concerned with how intelligence agents take actions in a given environment to maximize the cumulative reward they receive. In healthcare, applying RL algorithms could assist patients in improving their health status. In ride-sharing platforms, applying RL algorithms could increase drivers' income and customer satisfaction. For large language models, applying RL algorithms could align their outputs with human preferences. Over the past decade, RL has been arguably one of the most vibrant research frontiers in machine learning. Nevertheless, statistics as a field, as opposed to computer science, has only recently begun to engage with RL both in depth and in breadth. This chapter presents a selective review of statistical inferential tools for RL, covering both hypothesis testing and confidence interval construction. Our goal is to highlight the value of statistical inference in RL for both the statistics and machine learning communities, and to promote the broader application of classical statistical inference tools in this vibrant area of research.
\end{abstract}

\section{Introduction}
\label{s:intro}

Reinforcement learning (RL) is a trending research topic in machine learning that studies how artificial intelligence (AI) agents can make real-time decisions to maximize the long-term reward for human decision makers \citep{sutton2018reinforcement}. Over the past decade, it has been one of the most popular research directions in machine learning and AI. Google Scholar shows that in 2024, over 513,000 articles containing the keyword ``reinforcement learning" were published. At the 2024 AI top conference, ICML, 255 out of 2609 accepted papers studied RL, accounting for 10\% of the total accepted papers. 

There is a vast literature on RL, with numerous algorithms developed over the past few decades. From an application perspective, these algorithms can be categorized into online and offline learning, depending on whether they rely on real-time interactions with the environment or leverage pre-collected datasets \citep{levine2020offline}. From a technical perspective, they can be divided into model-based and model-free approaches, depending on whether they explicitly train a dynamic model for learning \citep{jiang2024note}. Within the model-free category, algorithms can be further classified into value-based and policy-based methods. Additionally, depending on the task objectives -- whether the goal is to learn an optimal policy or to evaluate the impact of a newly developed policy -- they can be categorized into policy optimization and policy evaluation \citep{dudik2014doubly}. 

Moreover, these algorithms have been successfully applied across a wide range of domains, including video games \citep{mnih2015human,openai2019dota}, the game of Go \citep{silver2016mastering}, mobile health \citep{liao2020personalized}, ride-sharing \citep{qin2025reinforcement}, and more recently, the development of large language models \citep{ouyang2022training}. For readers interested in RL algorithms, the aforementioned papers, along with the books by \citet{sutton2018reinforcement} and \citet{agarwal2019reinforcement}, as well as DeepMind and UCL’s RL lecture series (available on YouTube\footnote{\url{https://www.youtube.com/playlist?list=PLqYmG7hTraZDM-OYHWgPebj2MfCFzFObQ}}) and other publicly available lecture materials\footnote{see e.g., \url{https://github.com/callmespring/RL-short-course}}, may serve as valuable resources. 

In the statistics literature, RL is closely related to a vast body of work on estimating dynamic treatment regimes (DTRs), which aims to identify personalized treatment strategies for individual patients to maximize their health outcomes; see \citet{chakraborty2013statistical,kosorok2015adaptive,kosorok2019precision,tsiatis2019dynamic,li2023optimal} for reviews. However, most methods employ the so-called non-Markov decision process \citep{kallus2020double} -- where rewards and transitions depend on the entire data history -- to model the observed data. They often prove ineffective when applied to Markov decision processes \citep{kallus2022efficiently} -- which are more commonly studied in machine learning to formulate numerous sequential decision making problems. More recent work has shifted its focus to studying Markov decision processes for both policy optimization \citep{ertefaie2018constructing,luckett2020estimating,liao2022batch,yang2022toward,chen2024reinforcement,li2024settling,shi2024statistically,zhou2024estimating} and policy evaluation \citep{liao2021off,hu2023off,ramprasad2023online,wang2023projected,shi2024off}.


This paper will \underline{\textit{not}} elaborate on the aforementioned algorithms. Instead, it explores RL from a novel perspective. Specifically, we will introduce the application of traditional statistical inference methods, such as hypothesis testing and confidence interval construction \citep[see e.g.,][]{casella2024statistical}, in RL. We note that this topic has been less explored in the current RL literature. The goal of this review is to demonstrate the usefulness of statistical inference in RL to both the statistics and machine learning communities, and to encourage broader applications of statistical tools in this vibrant research area. 

The rest of the paper is organized as follows. Section \ref{sec:md} introduces some background concepts in RL, including the data generating process, the notion of policies, and the Markov decision process (MDP) model. Section \ref{sec:testMA} presents a case study and simulation experiments to illustrate the motivation and benefits of hypothesis testing in RL. Section \ref{sec:method} reviews existing tests for the Markov assumption underlying the MDP model, along with other relevant tests designed for conditional independence and the Markov assumption in time series analysis. Section \ref{sec:CI} provides a review of methodologies for off-policy confidence interval estimation, which aims to construct a confidence interval for a target policy's expected return using an offline dataset. Finally, Section \ref{sec:discuss} concludes the paper by discussing other related statistical inference topics.
\vspace{-8pt}

\section{Background: Data, Policies, and Models in Reinforcement Learning}\label{sec:md}

\begin{figure}[t]
    \centering
    \includegraphics[width=0.6\linewidth]{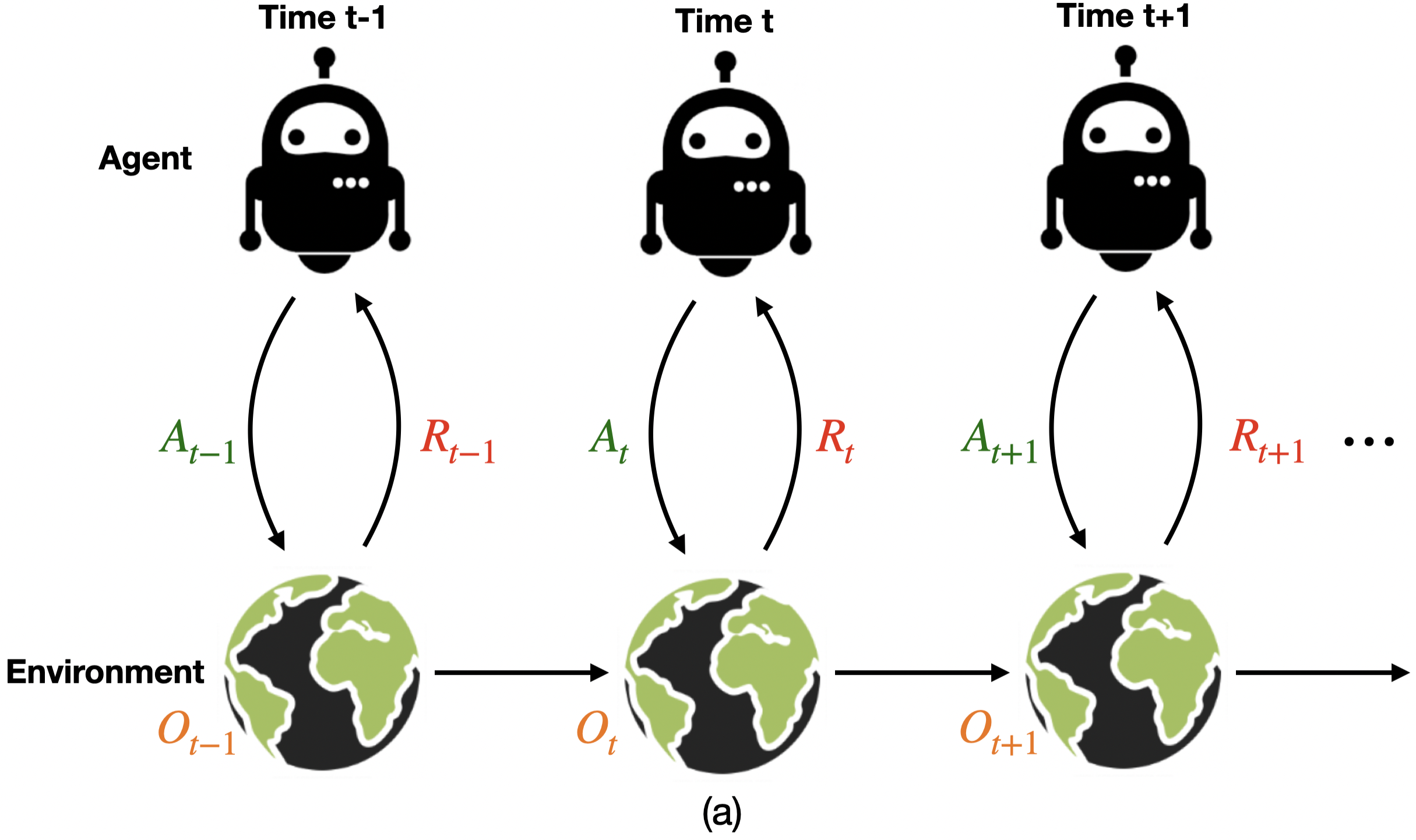}
    \includegraphics[width=0.3\linewidth]{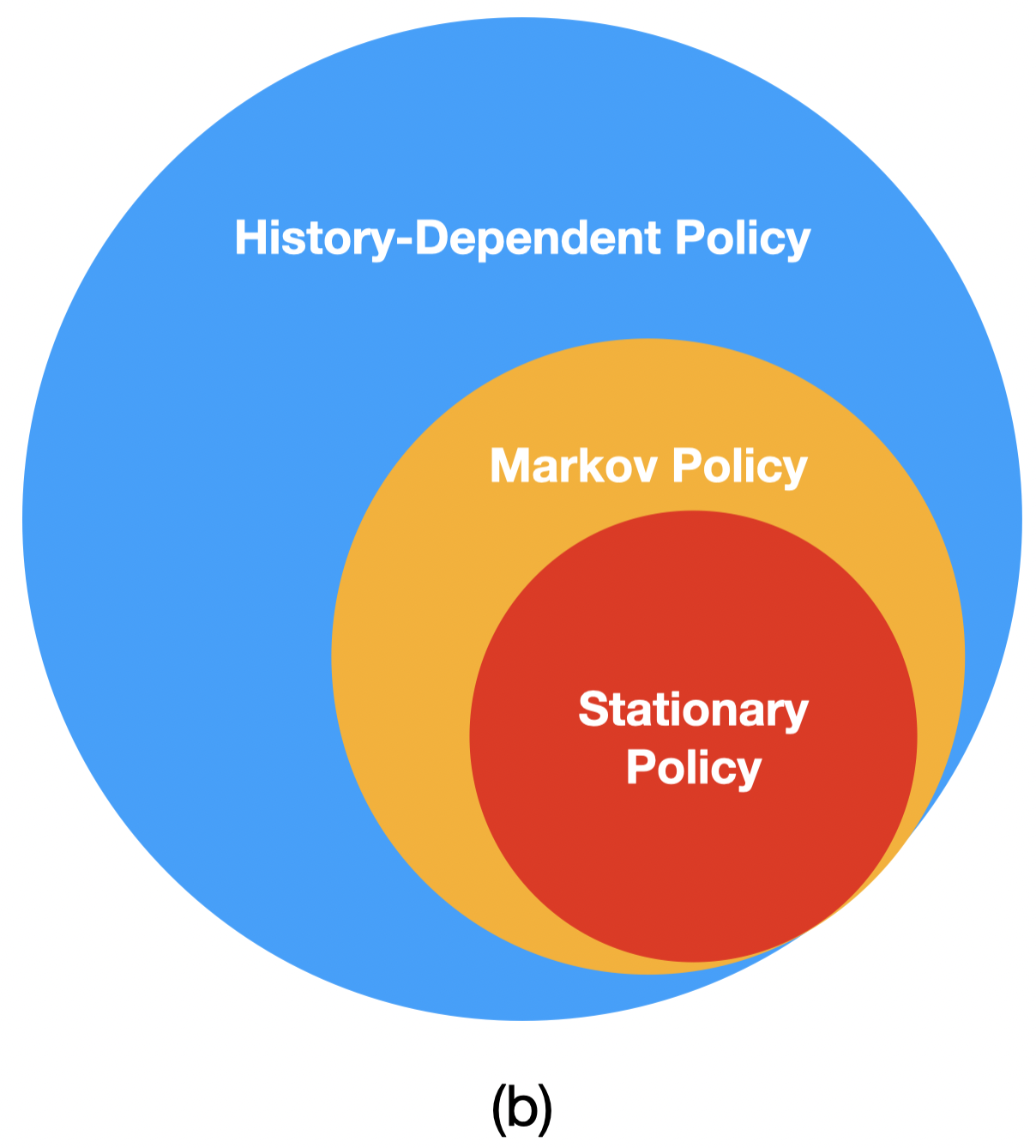}
    \caption{Visualizations of (a) sequential decision making; (b) three types of policies.}\label{fig:1}
\end{figure}

\textit{\textbf{Data}}: RL is concerned with solving the following sequential decision making problem:
\begin{itemize}[leftmargin=*]
    \item At each time step $t$, an agent observes certain features of the environment, denoted as $\Ot$;
    \item Based on \( \Ot \), the agent takes an action, denoted as \( \At \);
    \item The environment provides a corresponding reward, denoted as \( \Rt \), and transitions to a new state at the next time step.
\end{itemize}
This process repeats iteratively, as illustrated in Figure \ref{fig:1}(a). Thus, the observed data can be summarized into a sequence of ``observation-action-reward'' triplets \( (\Ot, \At, \Rt)_{t \geq 0} \). It is worth noting that the observation \( \Ot \) at each time step is not equivalent to the environment's state \( \St \). Indeed, the state can be viewed as a special observation with the Markov property, and we will elaborate on the difference between the two later. 

\textit{\textbf{Policies}}: The goal of RL is to learn an optimal policy $\pi^*$ based on the observation-action-reward triplets to maximize the agent's cumulative reward. Mathematically, a policy is defined as a conditional probability distribution function mapping the agent's observed data history to the action space. It specifies the probability of the agent taking different actions at each time step. Below, we introduce three types of policies (see Figure \ref{fig:1}(b) for a visualization of their relationships):
\begin{enumerate}[leftmargin=*]
    \item \textbf{History-dependent policy}: This is the most general form of policy. At each time \( t \), we define \( \bm{H}_t \) as the set containing the current observation \( \Ot \) and all prior historical information \( (\Oi, \Ai, \Ri)_{i < t} \). Under this policy, the selection of \( \At \) depends on the entire history \( \bm{H}_t \).
    \item \textbf{Markov policy}: The second type is a subset of the first. Under this policy, the conditional distribution of \( \At \) depends only on the current observation \( \Ot \), rather than the entire history \( \bm{H}_t \). In other words, for any \( t \) and \( a \), we have \( \mathbb{P}(\At = a | \bm{H}_t) = \mathbb{P}(\At = a | \Ot) \).
    \item \textbf{Stationary policy}: The third type is a subset of the second. Under this policy, the conditional distribution of \( \At \) not only satisfies the Markov property but also remains invariant over time. That is, for any \( t \), \( a \) and \( o \), we have \( \mathbb{P}(\At = a | \bm{H}_t) = \mathbb{P}(\At = a | \Ot) \) and \( \mathbb{P}(\At = a | \Ot = o) = \mathbb{P}(\Aini = a | \Oini = o) \).
\end{enumerate}
Following a given policy $\pi$, the expected cumulative reward the agent receives is given by $J(\pi)=\sum_{t\ge 0}\gamma^t \mathbb{E}^\pi (\Rt)$ where $0<\gamma<1$ denotes the discount factor that balances immediate and future rewards, and the expectation $\mathbb{E}^{\pi}$ is taken by selecting all actions according to $\pi$. The optimal policy $\pi^*$ is defined to be the maximizer of $J(\pi)$, i.e., $\pi^*=\arg\max_{\pi} J(\pi)$. 

\textit{\textbf{Models}}: Most RL algorithms employ the MDP \citep{puterman2014markov} to model the data. This model relies on two core assumptions:
\begin{enumerate}[leftmargin=*]
    \item \textbf{Markov assumption}: At any time step \( t \), given the current observation-action pair \( (\Ot, \At) \), the current reward \( \Rt \) and the next observation \( \Otn \) are conditionally independent of the prior history \( (\Oi, \Ai, \Ri)_{i < t} \).
    \item \textbf{Stationarity assumption}: For any observation \( o \) and action \( a \), the conditional distribution of \( \Rt \) and \( \Otn \) given \( \Ot = o \) and \( \At = a \) does not depend on time \( t \).
\end{enumerate}

These two assumptions provide a solid mathematical foundation for RL. Under these assumptions, it can be proven that there exists an optimal \underline{\textit{stationary policy}} $\pi^*$ whose expected return $J(\pi^*)$ is no smaller than that of any history-dependent policy \citep{puterman2014markov,ljungqvist2018recursive}. This result significantly reduces the computational complexity of RL, as it implies that we only need to search for the optimal policy within the set of stationary policies, rather than the much broader set of history-dependent policies. 

Despite decades of research in RL, methods for testing the validity of these assumptions remain underexplored. The following sections aim to provide a review of the available approaches, demonstrating how statistical inference can enhance the performance of modern RL algorithms. 
\vspace{-12pt}

\begin{figure}[t]
\centering
\includegraphics[width=0.7\linewidth]{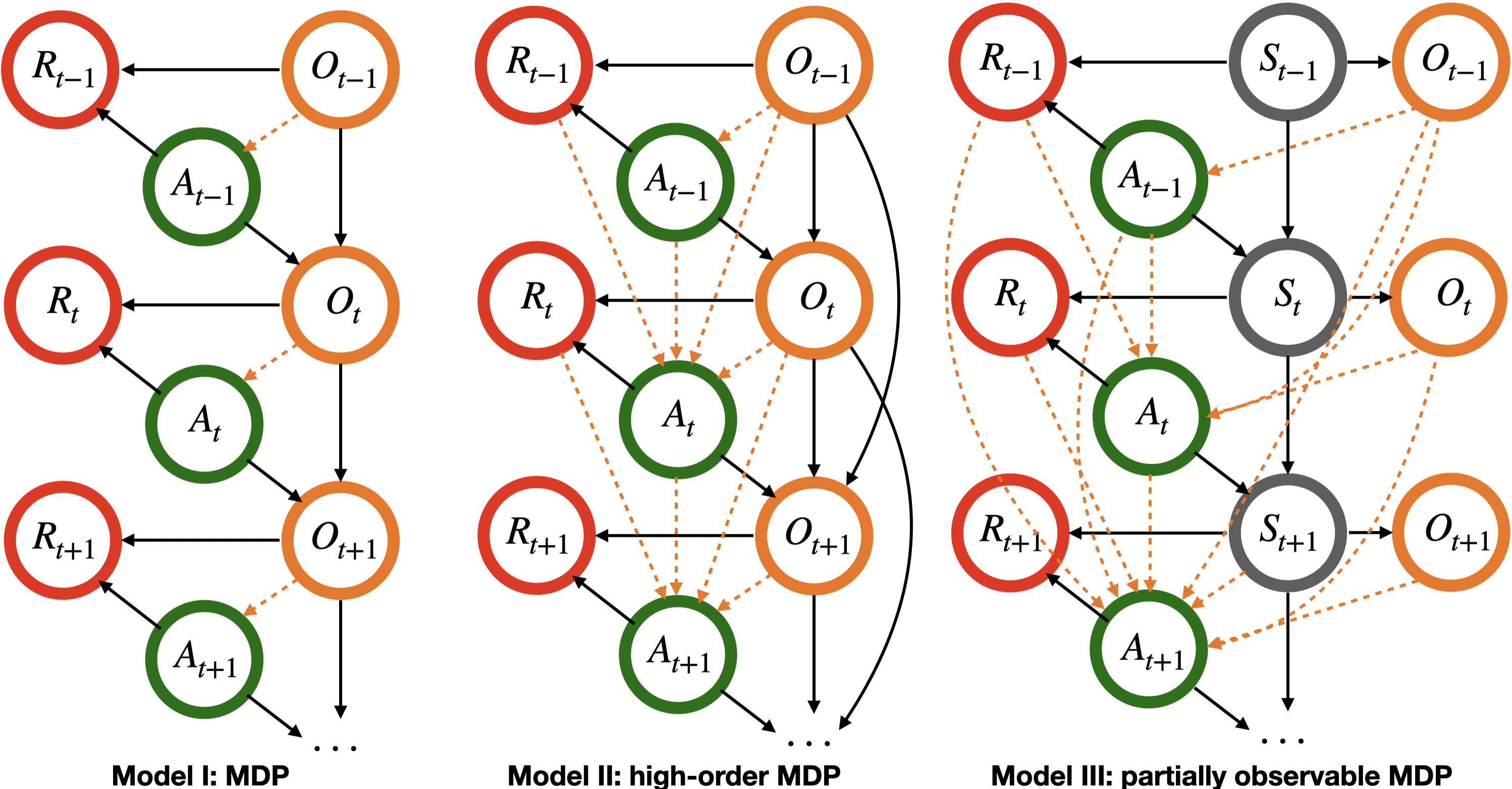}
\caption{Three RL models. Solid black lines represent causal relationships between variables, while dashed orange lines indicate the historical variables on which the optimal policy $\pi^*$ depends.}
\label{fig:model}
\end{figure}
\section{Hypothesis Testing in RL: Simulations and Case Studies}\label{sec:testMA}
We begin by discussing the necessity of testing the Markov assumption in RL. This paper primarily focuses on the offline setting where optimal policies and expected returns are estimated solely based on a pre-collected offline dataset without real-time interaction with the environment. The effectiveness of existing RL algorithms heavily relies on the Markov property of the data. To illustrate this, consider the three models depicted in Figure \ref{fig:model}:
\begin{enumerate}[leftmargin=*]
\item \textbf{MDP} (left panel): As described in Section \ref{sec:md}, the offline data satisfies the Markov property under the MDP model, and existing RL algorithms are directly applicable to estimate the optimal policy.

\item \textbf{High-order MDP} (middle panel): In this model, the offline data does not satisfy the first-order Markov assumption but exhibits a high-order Markov property. Specifically, $\Rt$ and $\Otn$ depend not only on $\At$ and $\Ot$, but also on the past $k\ge 2$ lagged variables $(\Oi,\Ai,\Ri)_{t-k<i<t}$. To apply existing RL algorithms, we merge the current observation $\Ot$ with these lagged variables to define a new state $\St$, so as to ensure the Markov property. The RL algorithm can then be applied to the transformed data triplets $(\St,\At,\Rt)_t$ to compute the optimal policy.

\item \textbf{Partially observable MDP} (POMDP, right panel): In this model, only a portion of the state or a noisy measurement of the state is observable. Even after merging current observations with a fixed number of preceding past lagged variables, the transformed data may not satisfy the Markov property. In the literature, specialized algorithms have been developed for policy learning in POMDPs \citep{krishnamurthy2016partially} and should be used to estimate the optimal policy if the offline data is known to originate from a POMDP.
\end{enumerate}
In practice, without prior knowledge, it is often unclear which model best describes a given offline dataset. If an underfitted model is chosen (e.g., assuming a second-order MDP when the true model is a third-order MDP), then the estimated policy may be suboptimal. Conversely, employing an overfitted model may yield consistent policies in large samples but suffers from increased variance in small samples due to the inclusion of many irrelevant lagged variables. Hypothesis testing is thus crucial for identifying the correct model type and determining its order:
\begin{itemize}[leftmargin=*]
    \item \textbf{Determining orders of MDPs}: If the data originates from a high-order MDP, hypothesis testing can help identify the correct order, allowing us to merge the appropriate lagged variables to define the state. We will illustrate this using a diabetes dataset in Section \ref{sec:casestudy}.
    \item \textbf{Identifying POMDPs}: If the data originates from a POMDP, hypothesis testing can guide us to use specialized algorithms tailored for policy optimization in POMDPs. We will illustrate this through simulation experiments in Section \ref{sec:tiger}.
\end{itemize}
\subsection{Case study: Diabetes datasets}\label{sec:casestudy}
Our analysis is based on the publicly available OhioT1DM dataset \citep{marling2018ohiot1dm}, accessible via this \href{https://webpages.charlotte.edu/rbunescu/data/ohiot1dm/OhioT1DM-dataset.html}{link}. The dataset includes eight weeks of data from six type-I diabetes patients\footnote{This dataset has been updated to include 12 patients \citep[see][for details]{marling2020ohiot1dm}, but we will use the original dataset with 6 patients for illustration.}. The goal is to use RL to estimate an optimal treatment policy that determines the insulin dosage for each patient at each time step to maintain stable blood glucose levels.

We discretize the eight-week data into hourly intervals, resulting in over 1,000 sample points per patient. To apply RL, we define the following observation, action and reward: 
\begin{itemize}[leftmargin=*]
    \item \textbf{Observation} is a three-dimensional vector including the patient’s average blood glucose level, calorie intake, and exercise intensity over the past hour. Here, blood glucose is our primary focus, while calorie intake and exercise intensity have a strong impact on it.
    \item \textbf{Action} corresponds to the insulin dosage administered during each hour, discretized into four levels, with 0 representing no insulin.
    \item \textbf{Reward} is determined by the blood glucose index \citep{rodbard2009interpretation}. A reward of 0 is given if the blood glucose level remains within the normal range; otherwise, the reward is negative, reflecting undesirable health outcomes.
\end{itemize}

We apply the hypothesis tests developed by \citet{shi2020does} and \citet{zhou2023testing}\footnote{Code available at: \href{https://github.com/RunzheStat/TestMDP}{TestMDP} and \href{https://github.com/yunzhe-zhou/markov_test}{markov\_test}.} to test whether the data follows a $k$-th order MDP for $k=1,2,\cdots,10$. Both tests assess the Markov assumption in RL. We will detail their methodologies in Section \ref{sec:method}. 
The corresponding $p$-values are listed in Table \ref{tab:realdata}. Small $p$-values indicate that the data does not conform to the $k$-th order MDP, whereas large $p$-values suggest that the null hypothesis that the data originates from a $k$-th order MDP cannot be rejected.

\begin{table}[t]
\centering
\caption{$p$-values and expected returns of estimated optimal policies for the diabetes dataset, under MDP models across different orders, ranging from $k=1$ to $10$.}\label{tab:realdata}
\footnotesize
\begin{tabular}{|l|c|c|c|c|c|c|c|c|c|c|}\hline
Order $k$ & 1 & 2 & 3 & 4 & 5 & 6 & 7 & 8 & 9 & 10\\ \hline
\citet{zhou2023testing}'s $p$-value & 0 & 0.010 & 0.030 & 0.240 & 0.243 & 0.421 & 0.436 & 0.485 & 0.360 & 0.338 \\ \hline
\citet{shi2020does}'s $p$-value & 0 & 0.001& 0.003& 0.097& 0.084& 0.092& 0.066& 0.069& 0.091& 0.103\\ \hline 
Expected return $J(\pi)$& -90.82 & -57.53 & -63.77 & -52.57 & -56.23 & -60.05 & -63.70 & -54.85 & -65.08 & -59.59 \\ \hline
\end{tabular}
\end{table}

At a significance level of $\alpha=0.05$, both tests by \citet{shi2020does} and \citet{zhou2023testing} reject the null hypothesis for $k<4$ but do not reject it for $k\ge 4$. These results imply that the data likely comes from a 4-th order MDP, meaning that data collected from four hours earlier is conditionally independent of future observations, given the data from the current four hours. Additionally, it can be seen that the test by \citet{shi2020does} generally yields smaller $p$-values than those obtained from the test of \citet{zhou2023testing}, despite that both tests are consistent in theory. This discrepancy arises because they employ different methods to estimate the Markov transition function; see Section \ref{sec:method} for details. According to simulation results in \citet{zhou2023testing}, the test by \citet{shi2020does} sometimes suffers from inflated type-I errors. Consequently, $p$-values from \citet{zhou2023testing} are more reliable. 

To further demonstrate the utility of these tests in improving existing RL algorithms, we conduct the following cross-validation procedure: (i) Split the data into training and testing data subsets, each containing three patients' data trajectories; (ii) Apply the fitted Q-iteration algorithm \citep[FQI,][]{ernst2005tree,riedmiller2005neural} to the training dataset to estimate the optimal policy under the $k$-th order MDP assumption; (iii) Apply the fitted Q-evaluation algorithm \citep[FQE,][]{le2019batch} to the testing dataset to estimate the expected return of the resulting policy; (iv) Repeat this process across all train-test splits, average the results and report them in Table \ref{tab:realdata}. The results clearly show that the policy estimated under the 4th-order MDP model assumption achieves the largest expected return, highlighting the importance of testing the Markov assumption in high-order MDPs. In particular, the Markov test can considerably enhance the performance of RL algorithms by enabling accurate determination of the MDP's order.

\subsection{Simulations: Tiger problem}\label{sec:tiger}
\begin{figure}[t]
    \centering
    \includegraphics[width=0.43\linewidth]{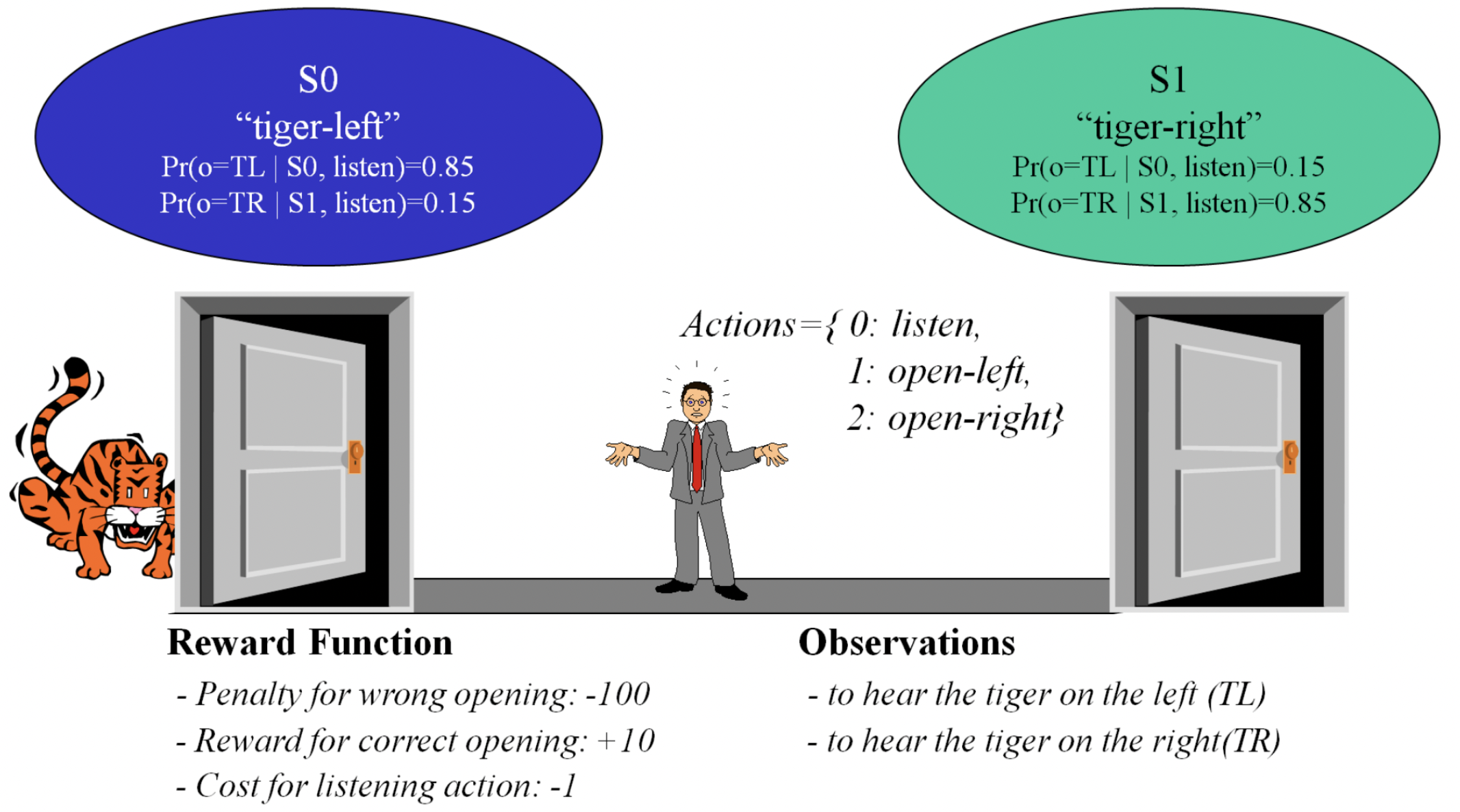}
    \includegraphics[width=0.25\linewidth,height=3.75cm]{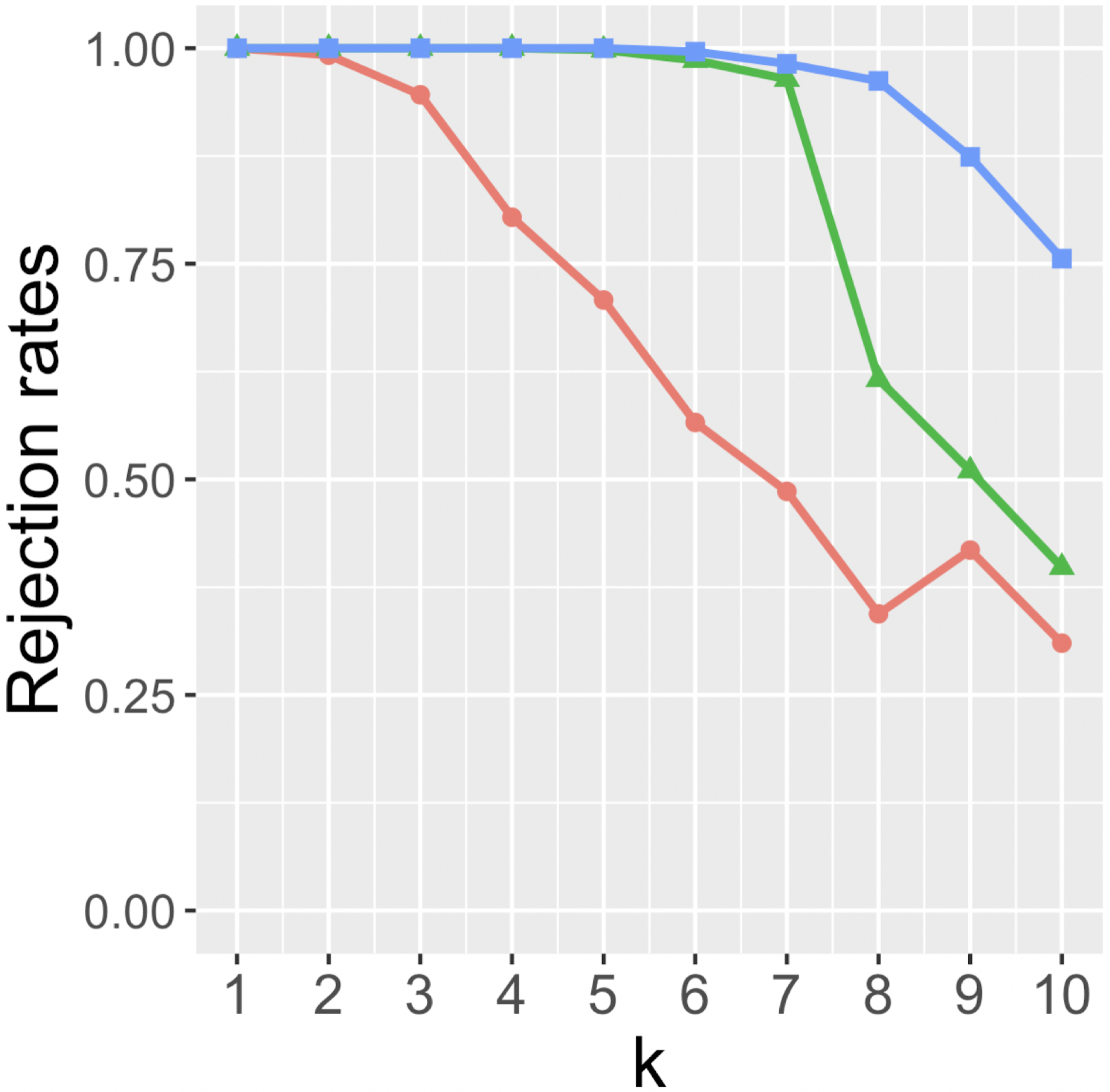}
    \includegraphics[width=0.3\linewidth,height=3.75cm]{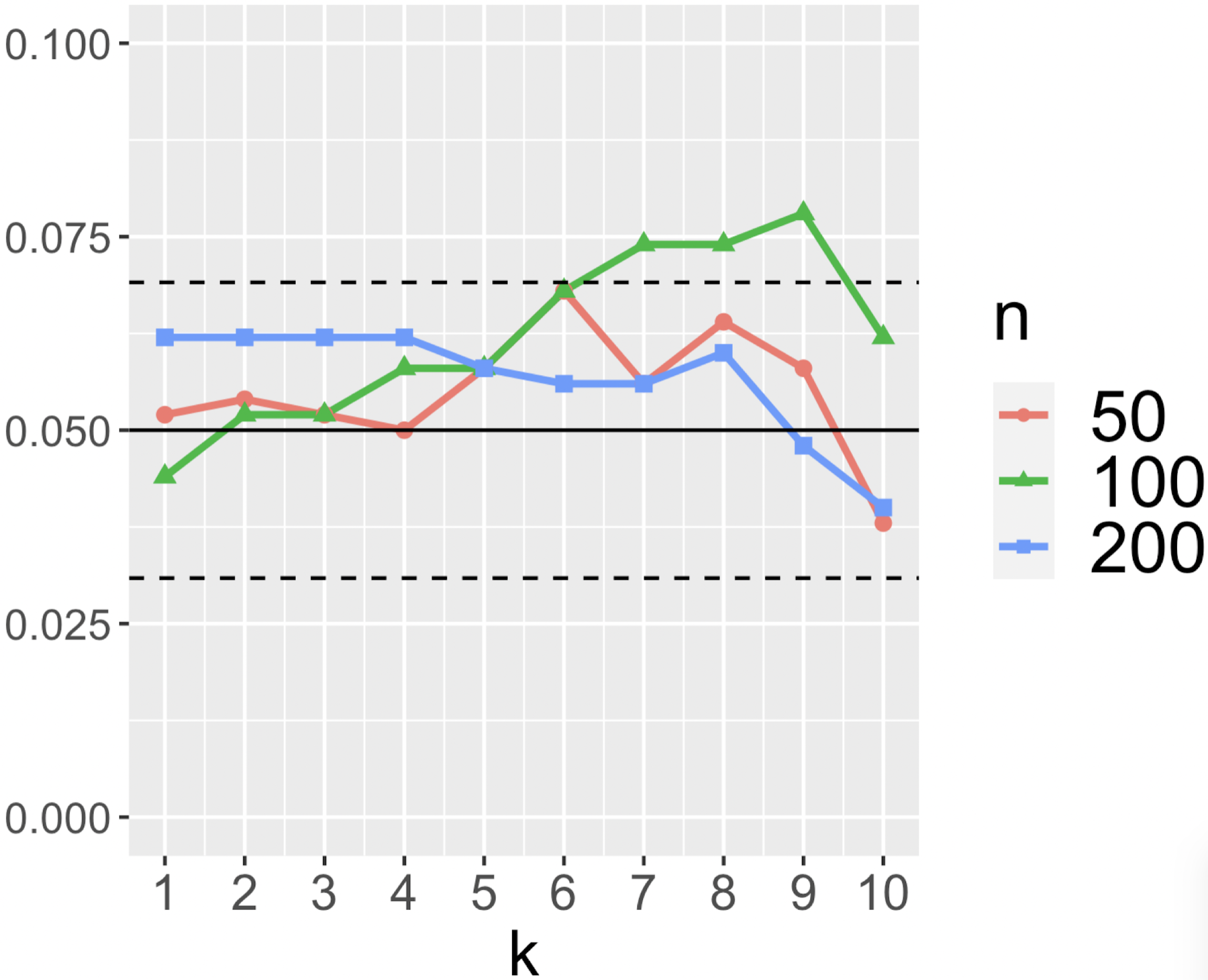}
    \caption{Left: The tiger problem. Middle: Percentage of rejections of the null hypothesis that the data from the tiger problem satisfies a $k$-th order Markov assumption using \citet{shi2020does}'s test, for $k=1,2,\cdots,10$. $n$ denotes the number of trajectories. Right: Same as Middle, but with the true location of the tiger known and included in the observations.}\label{fig:tiger}
\end{figure}

The simulation study is conducted using a classic POMDP environment: the tiger problem \citep{cassandra1994acting}. This environment can be described as follows (see the left panel of Figure \ref{fig:tiger}):
\begin{itemize}[leftmargin=*]
\item \textbf{State}: A tiger is randomly placed behind one of two doors at the initial time point.
\item \textbf{Action}: At each time step, the agent can (i) open the left door, (ii) open the right door, or (iii) listen to infer the tiger's location.
\item \textbf{Observation}: Listening provides an estimator of the tiger’s location, but with a 15\% probability of error.
\item \textbf{Reward}: (i) -100 for opening the door with the tiger, (ii) +10 for opening the empty door, and (iii) -1 for listening.
\end{itemize}
In this environment, the exact location of the tiger is a hidden state, and we cannot directly observe it. It can only be inferred through listening. Moreover, no matter how many times we listen, we cannot determine the true location of the tiger with 100\% certainty. Therefore, this environment does not satisfy the classic MDP or higher-order MDP assumption, but instead corresponds to a POMDP.

\citet{shi2020does} applied their test to this environment to test whether the generated data satisfies a  $k$-th order MDP, where 
$k=1,2,\cdots,10$. The significance level was set to 0.05, and we plot the proportion of rejections of the null hypothesis in 500 random simulations in the middle panel of Figure \ref{fig:tiger}. Since the true model follows a POMDP, the null hypothesis should theoretically be rejected regardless of the value of $k$. Thus, the proportions in the middle panel reflect the power of their test. It can be seen that the power increases as the number of trajectories $n$ grows and is very close to $1$ when $n\ge 100$ and $k\le 7$. This indicates that the test can effectively distinguish POMDPs from (higher-order) MDPs.

Additionally, \citet{shi2020does} conducted another simulation by including the true location of the tiger in the observation. In this case, the environment satisfies the MDP assumption due to the full observability of the state, and the null hypothesis holds regardless of the value of $k$. We also plot the proportion of rejections of the null hypothesis -- which corresponds to the type-I error -- in the right panel of Figure \ref{fig:tiger}, with upper and lower horizontal lines depicting the Monte Carlo error bounds. The results show that most proportions lie within the error bounds, demonstrating that the test can effectively identify (high-order) MDPs.

\section{Methodology: Forward-backward (Generative) Learning}\label{sec:method}
\textit{\textbf{Challenge}}. In this section, we focus on reviewing the hypothesis tests proposed by \citet{shi2020does} and \citet{zhou2023testing}, as both papers studied the RL setting\footnote{\citet{zhou2023testing} developed the test in the classic time series setting and later extended it to MDPs in Section 6.}. Later, we will discuss other relevant tests designed for conditional independence or the Markov assumption in time series analysis. Before delving into the details, we first outline the main challenges in testing the Markov property in RL and how these methods effectively address them. The primary challenge arises from the presence of high-dimensional continuous conditional variables. As we will discuss later, the Markov assumption is equivalent to a set of conditional independence conditions. Testing conditional independence is considerably more difficult than testing marginal independence when the conditional variables are continuous \citep{bergsma2004testing}. This challenge is further compounded when these conditional variables are high-dimensional. However, high-dimensional continuous variables are common in RL applications. For example, testing a high-order MDP requires to merge the current observation with past lagged variables to define the state, which naturally introduces high-dimensional conditional variables. In the diabetes study, variables such as blood glucose levels and calorie intake are 
continuous. 

\citet{shi2020does} and \citet{zhou2023testing} leveraged modern supervised learning and deep generative learning algorithms to handle high-dimensional continuous conditional variables, respectively. Take \citet{zhou2023testing}'s proposal as an example. The idea is to learn two generators: (i) \underline{\textit{a forward generator}}, which predicts the next observation based on the current observation-action pair, and (ii) 
\underline{\textit{a backward generator}}, which reconstructs the current observation and action based on the next observation-action pair. By combining these two generators, \citet{zhou2023testing} constructed a doubly robust test statistic, effectively eliminating the bias from a single generator and ensuring the test statistic has a desirable limiting distribution. \citet{shi2020does} constructed a similar doubly robust test statistic with \underline{\textit{a forward learner}} and \underline{\textit{a backward learner}}. We elaborate on these forward-backward-learning-based methods below.

\textit{\textbf{Null and alternative hypotheses}}. 
To simplify the discussion, we assume that the reward $\Rt$ is a deterministic function of the observation $\Ot$, action $\At$, and the next observation $\Otn$. This assumption holds automatically in the diabetes case study, where $\Rt$ depends only on the blood glucose level at time $t+1$, which is included in $\Otn$. To ensure this assumption is always satisfied, we can redefine the observation by incorporating $\Rt$ into $\Otn$. Based on this assumption, we consider the following null and alternative hypotheses:

\begin{itemize}[leftmargin=*]
\item $\mathcal{H}_0$: For any time $t$, given $\Ot$ and $\At$, $\Otn$ is conditionally independent of past observation-action pairs $(\Oi,\Ai)_{i<t}$.
\item $\mathcal{H}_a$: There exists at least some $t$ where $\Otn$ is not conditionally independent of past observation-action pairs given $\Ot$ and $\At$.
\end{itemize}

According to Theorem 1 in \citet{shi2020does} and \citet{zhou2023testing}, under the stationarity assumption, the null hypothesis is equivalent to the intersection of a series of conditional independence hypotheses $\cap_{q> 1} \mathcal{H}_0(q)$ where 
\begin{itemize}
    \item $\mathcal{H}_0(q)$: For any time $t$ and integer $q>1$, given $(\Oi,\Ai)_{t<i<t+q}$, $\Otq$ is conditionally independent of $(\Ot,\At)$.
\end{itemize}
Thus, testing the Markov property is equivalent to testing these conditional independence hypotheses, each of which assesses the conditional independence of observations separated by $q$ time steps.

\textit{\textbf{Test statistics}}. To elaborate the two test statistics, notice that under $\bm{\mathcal{H}}_0(q)$, we have for any functions $g$ and $h$, 
\begin{equation*}
    \mathbb{E} \{g(\Otq)-\mathbb{E}[g(\Otq)|(\Oi,\Ai)_{t<i<t+q}]\}\{h(\Ot,\At)-\mathbb{E}[h(\Ot,\At)|(\Oi,\Ai)_{t<i<t+q}]\}=0.
\end{equation*}
Using the Markov assumption, it is immediate to see that the above two conditional expectations depend only on    
$({\color{orange}\bm{O}_{t+q-1}},{\color{ao}\bm{A}_{t+q-1}})$ and $({\color{orange}\bm{O}_{t+1}},{\color{ao}\bm{A}_{t+1}})$, respectively. This leads to 
\begin{equation}\label{eqn:key}
    \mathbb{E} \{g(\Otq)-\mathbb{E}[g(\Otq)|({\color{orange}\bm{O}_{t+q-1}},{\color{ao}\bm{A}_{t+q-1}})]\}\{h(\Ot,\At)-\mathbb{E}[h(\Ot,\At)|\Otn,{\color{ao}\bm{A}_{t+1}} ]\}=0.
\end{equation}
Equation \eqref{eqn:key} is the key to the tests by \citet{shi2020does} and \citet{zhou2023testing}. Specifically, both estimate the two conditional expectations $\mathbb{E} [g(\Otn)|\Ot,\At]$ and $\mathbb{E} [h(\Ot,\At)|\Otn,\Atn]$ for a set of functions $g$ and $h$, plug-in these estimators into Equation \eqref{eqn:key}, and approximate the leftmost expectation using the sample average to construct the test statistic $S(q,g,h)$. Under the null hypothesis, $S(q,g,h)$ should be close to 0 for all $q$, $g$, and $h$. If there exist combinations of $q$, $g$, and $h$ for which the statistic $S(q,g,h)$ significantly deviates from 0, this provides evidence to reject the null hypothesis, indicating the violation of the Markov property. Based on this, \citet{shi2020does} and \citet{zhou2023testing} considered a large number of combinations of $q$, $g$, and $h$, and construct a maximum-type test statistic $\max_{q,g,h} |S(q,g,h)|$. If this test statistic exceeds a certain threshold (discussed below), the null is rejected.

We next provide some remarks on Equation \eqref{eqn:key} and discuss some implementation details regarding the construction of the test statistics:

\begin{itemize}[leftmargin=*]
\item According to Theorem 2 in \citet{shi2020does} and \citet{zhou2023testing}, under the null hypothesis, Equation \eqref{eqn:key} holds as long as either one of the two conditional expectations, $\mathbb{E} [g(\Otn)|\Ot,\At]$ or $\mathbb{E} [h(\Ot,\At)|\Otn,\Atn]$ is correctly specified. Such a  \underline{\textit{double robustness}} property greatly facilitates hypothesis testing. Specifically, when the convergence rates of these conditional expectations are $o_p(N^{-1/4})$ where $N$ denotes the total sample size (i.e., the total number of observation-action-next-observation triplets), the resulting test becomes consistent. We do not require them to achieve the standard parametric convergence rate; see Condition (C4) in \citet{shi2020does} and Condition 4 in \citet{zhou2023testing}.
\item Given the need to consider a large number of \( g \) and \( h \) combinations, it is essential to develop an efficient algorithm for computing the conditional expectations. \citet{shi2020does} proposed to employ random forests to model these conditional expectations and adapted the quantile random forest algorithm \citep{meinshausen2006quantile} to accelerate computation. Specifically, in their proposal, the structure of the forest (e.g., the tree depth and the split points) is independent of \( g \) (or \( h \)), meaning that the same forest is used for any \( g \) (or \( h \)). The final conditional expectation is computed based on the specific form of \( g \) (or \( h \)), which substantially simplifies the computation since the forest does not need to be retrained for different \( g \) (or \( h \)). For details, see Section 5.1 of \citet{shi2020does}.

\item Similarly, \citet{zhou2023testing} employ generative learning algorithms to improve the computational efficiency. Take the computation of \( \mathbb{E}[g(\Otn) | \Ot, \At] \) as an example. The key idea is to learn the conditional distribution of \( \Otn \) given \( \Ot \) and \( \At \). This requires to learn a generator that takes \( \Ot \), \( \At \), and random noise as inputs and outputs \( {\color{orange}\bm{O}^*} \), such that the distribution of \( {\color{orange}\bm{O}^*} \) closely approximates the conditional distribution of \( \Otn \). By generating \( M \) independent samples \( \{{\color{orange}\bm{O}^{(m)}}\}_{m=1}^M \) from this generator, the conditional expectation can be approximated as \( M^{-1} \sum_{m=1}^M g({\color{orange}\bm{O}^{(m)}}) \). In the experiments, \citet{zhou2023testing} used the mixture density network \citep{bishop1994mixture} to model the conditional distribution. Meanwhile, other generative neural network models, such as generative adversarial networks \citep[GANs,][]{goodfellow2014generative} and diffusion models \citep{ho2020denoising} are equally applicable; see e.g., \citet{bellot2019conditional}, \citet{shi2021double,shi2023testing} and \citet{zhang2024doubly,zhang2025testing} for their use in conditional (mean) independence testing. 

\item Both \citet{shi2020does} and \citet{zhou2023testing} set \( g \) and \( h \) to characteristic functions because they uniquely determine the distribution function. Alternatively, neural networks can also be used for $g$ and $h$, as suggested by \citet{shi2021double}.

\item Both methods employ cross-fitting to construct the test statistics. Specifically, the data is split into, say, $K$ non-overlapping subsets. One subset is used to estimate the conditional expectations, while the remaining $K-1$ subsets are used to construct the test statistic \( S(q, g, h) \). This process is repeated for different combinations of data subsets, switching their roles in estimating the conditional expectations and constructing the test statistic. The test statistics from all $K$ combinations are then averaged, and the maximum over \( q \), \( g \), and \( h \) is taken. We remark that cross-fitting eliminates the dependence between the data used for constructing the statistic and that used for estimating conditional expectations, ensuring consistency without requiring VC-class-type conditions on the conditional expectation estimators. This technique is commonly used in the construction of doubly robust estimators \citep{chernozhukov2018double,kallus2022efficiently}.
\end{itemize}

\textbf{\textit{Rejection region}}. After computing the test statistic \(\max_{q,g,h}|S(q,g,h)|\), we need to determine its critical value to decide whether to reject the null hypothesis (depending on whether the test statistic exceeds this critical value). Here, we employ a high-dimensional bootstrap method \citep{chernozhukov2014gaussian} to estimate the critical value. Although \citet{chernozhukov2014gaussian} assumed independent data, their method is also applicable to time series and MDPs under the Markov assumption, as each $S(q,g,h)$ forms a sum of martingale differences, exhibiting properties similar to independent sums \citep{belloni2018high}. See also \citet{chernozhukov2023high} for a review of these high-dimensional bootstrap methods. 

Specifically, under the null hypothesis, all $S(q,g,h)$ asymptotically follow a multivariate Gaussian distribution. Thus, $\max_{q,g,h} |S(q,g,h)|$ converges in distribution to the maximum of several multivariate Gaussian variables. We first estimate the covariance matrix of these statistics using their sample covariance. Then, we simulate random vectors from a Gaussian distribution with the same covariance matrix, take the absolute value of each component, and compute the maximum. By repeating this process, we obtain a series of bootstrap statistics. 
Finally, the threshold is set to the upper $\alpha$-quantile of these bootstrap statistics. If the test statistic exceeds this threshold, the null hypothesis is rejected. Theoretical properties of this test are detailed in Theorem 1 of \citet{shi2020does} and Theorems 5 and 6 of \citet{zhou2023testing}.

\textit{\textbf{Comparison}}. To better understand the aforementioned forward-backward-learning-based tests, we compare them with existing methods for testing conditional independence and the Markov property:
\begin{itemize}[leftmargin=*]
\item \textbf{Single-regression/generative-learning-based tests}: \citet{chen2012testing} and \citet{bellot2019conditional} proposed tests for the conditional independence condition and the Markov assumption in time series analysis, respectively. Both methods learned a single generator/regression function, which results in test statistics that are not doubly robust and can suffer from large biases. Specifically:
\begin{itemize}
\item \citet{bellot2019conditional} used GANs to learn the generator, but their method fails to control the type-I error unless unless applied in a semi-supervised learning setting with a large amount of unlabeled data \citep{berrett2020conditional}.
\item \citet{chen2012testing} used local polynomial regression to estimate the conditional expectation. They employed undersmoothing -- which reduces the bias of the test statistic at the cost of increased variance -- to theoretically control the type-I error. However, in practice, their test often fails to achieve consistency in moderate to high-dimensional settings; see the simulation results in \citet{zhou2023testing} for details.
\end{itemize}

\item \textbf{Double-regression-based tests}: \citet{zhang2019measuring}, \citet{shah2020hardness} and \citet{quinzan2023drcfs} proposed double-regression-based tests for conditional independence. Their test statistics are doubly robust as well. While conceptually similar to \citet{shi2020does} and \citet{zhou2023testing}, these methods focus solely on conditional independence instead of the Markov assumption. Moreover, their test functions \( g \) and \( h \) are limited to identity functions, which simplifies the testing but reduces its power; see \citet{shi2021double} and \citet{shi2023testing} for empirical comparisons with these double-regression-based tests in the context of conditional independence testing.

\item \textbf{Strong conditional independence tests}: The proposed forward-backward-learning-based tests primarily focus on weak conditional independence \citep{daudin1980partial} and cannot detect all alternative hypotheses. In other words, there exist alternative hypotheses under which the left-hand-side of Equation \eqref{eqn:key} is 0 for all \( q \), \( g \), and \( h \). To address this, strong conditional independence tests can be developed to detect all alternatives; see the discussion following Theorem 5 in \citet{shi2021double}. However, while weak conditional independence tests may fail to detect certain alternatives, they often have larger power in most practical scenarios.
\end{itemize}

\textbf{\textit{Model selection}}. To conclude this section, we discuss model selection based on the forward-backward-learning-based tests. The method is similar to forward selection in regression analysis: Given an offline dataset and a significance level, we first test whether the data satisfies the Markov assumption. If the test is not rejected, we conclude that the data follows an MDP. If rejected, we sequentially test whether the data follows a second-order MDP, a third-order MDP, and so on, until the null hypothesis is not rejected at some order $k$ and we conclude that the data follows a $k$-th order MDP. Finally, if the null hypothesis is rejected at a very large order, then the data is more likely to follow a POMDP. We also note a recent proposal by \citet{ye2024consistent}, which directly addresses order selection in MDPs without relying on hypothesis testing.

\begin{table}[t]
\centering
\caption{A summary of existing methodologies for off-policy confidence interval estimation.}\label{tab:OPCIE}
\footnotesize
\begin{tabular}{|c|c|c|c|c|}\hline
& Model-based & Direct method & Importance sampling & Doubly robust \\ \hline
Concentration & & \multirow{3}*{\citet{feng2020accountable}} & \multirow{3}*{\citet{thomas2015high}} & \citet{thomas2016data} \\ 
inequalities  & & & & \citet{jiang2016doubly} \\ 
 & & & & \citet{zhou2023distributional} \\ \hline
Normal & & \citet{luckett2020estimating} & \multirow{3}*{\citet{wang2023projected}} & \citet{shi2021deeply} \\ 
approximation & & \citet{liao2021off} & & \citet{liao2022batch} \\
& & \citet{shi2022statistical} & & \citet{kallus2022efficiently} \\ \hline
\multirow{2}*{Bootstrap} & \multirow{2}*{\citet{hanna2017bootstrapping}} & \multirow{2}*{\citet{hao2021bootstrapping}} & & \citet{thomas2016data} \\
& & & & \citet{hanna2017bootstrapping} \\ \hline
Empirical & & & \multirow{2}*{\citet{dai2020coindice}} & \\ 
likelihood & & & & \\ \hline
\end{tabular}
\end{table}

\section{Off-policy Confidence Interval Estimation}\label{sec:CI}
Off-policy evaluation (OPE) aims to assess the impact of implementing a newly developed target policy $\pi$ based on a historical dataset collected from a possibly different behavior policy. There is an extensive literature on OPE; see \citet{uehara2022review} for a recent review. OPE is particularly relevant to statistical inference, as in many applications, we are not only interested in the point estimator of the target policy $\pi$'s expected return $J(\pi)$, but also require uncertainty quantification for this estimator. 

We summarize the existing OPE methodologies in Table \ref{tab:OPCIE}, which can be broadly categorized into four groups based on the approach used to construct confidence intervals (CIs). Each group corresponds to a row in the table and reflects a specific type of methods for CI construction: concentration inequalities \citep{zhang2020concentration}, normal approximation \citep{casella2024statistical}, bootstrap \citep{efron1994introduction}, and empirical likelihood \citep{owen2001empirical}.

Meanwhile, these methodologies can also be classified according to the type of point estimator used to estimate $J(\pi)$, as summarized across the columns of the table.  depending on the type of point estimator used to estimate $J(\pi)$. Specifically, model-based method derives the estimated $J(\pi)$ by learning the underlying MDP model; the direct method learns $J(\pi)$ based on an estimated value or Q-function; Importance sampling (IS) reweight the immediate reward using IS ratios to address the distributional shift between target policy and behavior policy; and doubly robust (DR) method combines the direct method and IS for more robust estimation of $J(\pi)$. 

We use DR \citep[see e.g.,][]{zhang2013robust,jiang2016doubly,thomas2016data,uehara2020minimax,kallus2022efficiently,liao2022batch} as an example to elaborate on the CI construction. Under the Markov assumption, when the estimated IS ratio and Q-function converge at a rate faster than $O_p(N^{-1/4})$, the DR estimator is asymptotically equivalent to a sum of martingale differences, which behaves similarly to i.i.d. sums. As a result, the standard sampling variance estimator can be used to quantify their standard error, enabling the resulting Wald-type CI to achieve nominal coverage. When the IS ratio or Q-estimator fails to achieve this convergence rate, high-order influence functions can be employed to further reduce the bias of the resulting estimator and ensure the validity of the Wald-type CI \citep{shi2021deeply}.

To conclude this section, we remark that nearly all the aforementioned work assumed that the target policy $\pi$ is known in advance. In settings where 
$\pi$ corresponds to the optimal policy $\pi^*$, which needs to be estimated, constructing confidence intervals becomes particularly challenging in nonregular cases where $\pi^*$ is non-unique. Several methods have been proposed in the statistics literature to address this issue, including the 
$m$-out-of-$n$ bootstrap \citep{chakraborty2014inference}, the online one-step method \citep{luedtke2016statistical}, and subagging \citep{shi2020breaking}.

\section{Discussion}\label{sec:discuss}
In this section, we conclude the paper by presenting a review of some other related topics that lie in the intersection of statistical inference and RL:  
\begin{itemize}[leftmargin=*]
    \item \textbf{Testing the stationarity assumption}: This paper primarily focuses on hypothesis testing of the Markov assumption in RL, assuming stationarity of the observation-action-reward triplets over time. However, the stationarity assumption itself can be violated. In the literature, several studies have proposed tests for the stationarity assumption, covering both model-free \citep{padakandla2020reinforcement,li2025testing} and model-based approaches \citep{alegre2021minimum,wang2023robust}. Results from these studies demonstrate that such tests enable accurate change point detection in nonstationary environments, ultimately enhancing the performance of the subsequent policy optimization.
    \item \textbf{A/B testing}: A/B testing is widely used in technological industries for comparing the performance of a newly developed policy against a standard control during product deployment \citep[see][for reviews]{larsen2024statistical,quin2024b}. It is inherently a statistical inference problem that computes \( p \)-values to assess the significance of improvements achieved by the new policy relative to the standard one. Recently, there has been growing interest in integrating RL into A/B testing to evaluate long-term treatment effects, which are common in practice \citep{glynn2020adaptive,farias2022markovian,li2024optimal,shi2023dynamic,wen2024analysis}. This emerging research direction warrants further exploration.
    \item \textbf{Statistical inference in bandits}: Finally, bandit is a special class of RL problems, characterized by independent state transitions. There has been a growing literature focused on statistical inference for adaptively collected data using bandit algorithms \citep{zhang2020inference,bibaut2021post,dimakopoulou2021online,hadad2021confidence,simchi2023multi}.  
\end{itemize}

\bibliographystyle{plainnat}
\bibliography{bibtex_example}

\end{document}